\documentclass[letterpaper]{article} 
\usepackage{aaai24}  
\usepackage{times}  
\usepackage{helvet}  
\usepackage{courier}  
\usepackage[hyphens]{url}  
\usepackage{graphicx} 
\usepackage{natbib}  
\usepackage{caption} 
\usepackage{algorithm}
\usepackage{algorithmic}

\usepackage{newfloat}
\usepackage{listings}
\DeclareCaptionStyle{ruled}{labelfont=normalfont,labelsep=colon,strut=off} 
\floatstyle{ruled}
\newfloat{listing}{tb}{lst}{}
\floatname{listing}{Listing}
\usepackage{multirow}
\usepackage{amsmath}
\usepackage{amsfonts}

\title{Fast Machine Unlearning Without Retraining Through Selective Synaptic Dampening}
\author{
    Jack Foster\equalcontrib\textsuperscript{\rm 1,2},
    Stefan Schoepf\equalcontrib\textsuperscript{\rm 1},
    Alexandra Brintrup\textsuperscript{\rm 1,2}
}
\affiliations{
    \textsuperscript{\rm 1}University of Cambridge, Department of Engineering\\
    \textsuperscript{\rm 2}The Alan Turing Institute\\

    \{jwf40, ss2823, ab702\}@cam.ac.uk
}

\usepackage{bibentry}

\begin{document}

\maketitle

\begin{abstract}

Machine unlearning, the ability for a machine learning model to forget, is becoming increasingly important to comply with data privacy regulations, as well as to remove harmful, manipulated, or outdated information. The key challenge lies in forgetting specific information while protecting model performance on the remaining data.  While current state-of-the-art methods perform well, they typically require some level of retraining over the retained data, in order to protect or restore model performance. This adds computational overhead and mandates that the training data remain available and accessible, which may not be feasible. In contrast, other methods employ a retrain-free paradigm, however, these approaches are prohibitively computationally expensive and do not perform on par with their retrain-based counterparts. We present Selective Synaptic Dampening (SSD), a novel two-step, post hoc, retrain-free approach to machine unlearning which is fast, performant, and does not require long-term storage of the training data. First, SSD uses the Fisher information matrix of the training and forgetting data to select parameters that are disproportionately important to the forget set. Second, SSD induces forgetting by dampening these parameters proportional to their relative importance to the forget set with respect to the wider training data. We evaluate our method against several existing unlearning methods in a range of experiments using ResNet18 and Vision Transformer. Results show that the performance of SSD is competitive with retrain-based post hoc methods, demonstrating the viability of retrain-free post hoc unlearning approaches. 

\end{abstract}

\section{Introduction}


\begin{figure*}[]
    \centering
    \includegraphics[width=0.74\textwidth]{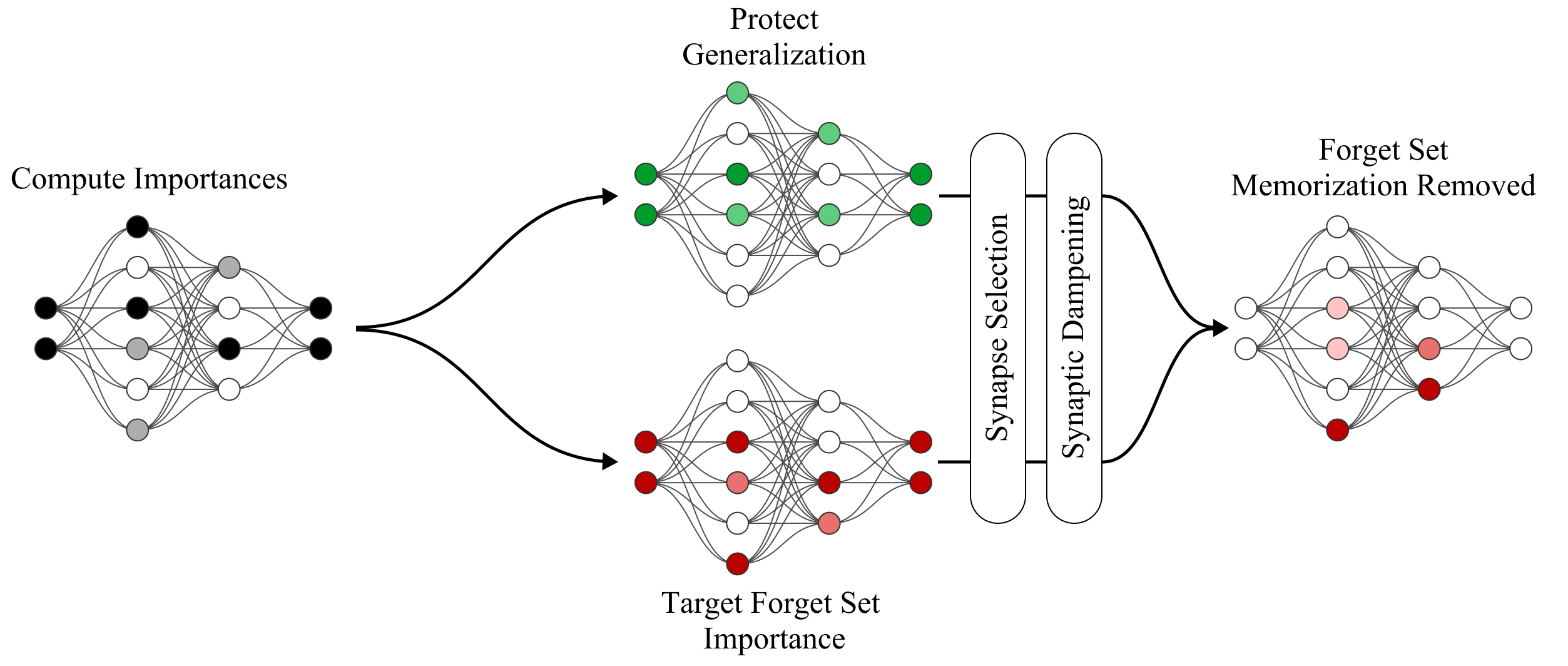}
    \caption{The Selective Synaptic Dampening process. Importance is calculated using the Fisher information. SSD identifies the parameters that are specialized towards the forget set, and dampens them proportional to this specialization}
    \label{fig:method}
\end{figure*}

Modern machine learning (ML) models are trained on vast amounts of data, much of which may be sensitive, private, or copyrighted. To address threats posed by large-scale data collection, authorities are enacting data privacy regulations that afford individuals the right to request the deletion of their data (e.g., GDPR \citep{voigt2017eu}). Despite the increasing need to facilitate forgetting, there is much work to be done in designing such algorithms. The process of forgetting information within an ML model is referred to as machine unlearning.

The challenge of machine unlearning can be thought of as a multi-objective task, conducting forgetting without degrading model performance on the remaining data. \citet{nguyen2022survey} refers to this trade-off in terms of design requirements, referring to performance preservation as keeping the model completeness, and unlearning efficiency as timeliness and light-weightiness. Timeliness is a key constraint, as full retraining of a model without the to-be-forgotten data would yield the desired results but doing so is time and resource-intensive. Similarly, light-weightiness refers to what preparation is necessary for the unlearning process, such as storing a list of samples and parameter updates for every training batch as in \citet{graves2021amnesiac}. This adds significant overhead and cannot be performed post-hoc.

Current state-of-the-art approaches rely on various retraining or fine-tuning steps in order to preserve model performance while unlearning the specified data \citep{tarun2023deep, tarun2023fast, chundawat2023can, chundawat2023zero, graves2021amnesiac}. This can add overhead and, importantly, mandates that the training data be stored permanently.

In this paper, we propose Selective Synaptic Dampening (SSD), a retraining-free, post hoc unlearning approach to enable lightweight and timely unlearning. We achieve this by distinguishing between generalized and specialized information, prioritising the protection of generalized, broadly useful information while dampening parameters that are specialized towards to-be-forgotten samples. SSD builds on the finding that overparameterised ML models are prone to memorization of training data \citep{lee2011long, feldman2020does, carlini2019secret}. Thus, we contend that targeting this specialized information can induce forgetting while minimising influence on the generalization capability of the model. The remaining information in the model is generalized and therefore not violating individual privacy (e.g., the ability to detect the shape of a person versus detecting Jane Smith). We use the diagonal of the Fisher information matrix (FIM) to identify these specialized parameters. \citet{golatkar2020eternal} have also proposed a retraining-free unlearning approach based on the FIM, but as shown by \citet{tarun2023fast} and our own benchmarks, their approach does not meet key design criteria. First, the computational effort exceeds retraining and takes orders of magnitude longer than the unlearning method of \citet{tarun2023fast} based on their benchmarks. Second, their benchmarks also show that the unlearning performance does not match current state-of-the-art methods. We address both of these shortcomings via Selective Synaptic Dampening. Fig. \ref{fig:method} provides a diagrammatic overview of the SSD process.

We benchmark SSD against state-of-the-art unlearning methods \citep{chundawat2023can, graves2021amnesiac, tarun2023fast} with three different unlearning scenarios: (i) single-class forgetting \citep{chundawat2023zero}, (ii) sub-class forgetting \citep{Golatkar_2020_CVPR, golatkar2020forgetting}, (iii) random observations forgetting \citep{golatkar2021mixed}.
Experimental results show that SSD is orders of magnitude faster than previous retrain-free methods \citep{golatkar2020eternal, golatkar2020forgetting} while performing comparably to established retrain-based methods both in terms of speed as well as forgetting performance.

We make the following key contributions:


\begin{enumerate}
    \item We propose a novel retraining-free selective unlearning method that is competitive with state-of-the-art retraining-based methods.
    \item We consider unlearning as a selective task in which only a small number of parameters should be modified to preserve model consistency. 
    \item SSD only needs access to the training data once to compute the FIM and can discard it afterwards, reducing storage requirements compared to retraining-based methods.
\end{enumerate}

\section{Related Work}
\label{sec:related}

\textbf{Differential privacy.} Differential privacy seeks to provide guarantees that information about individuals in a dataset is not leaked by the output of some model or function that uses this data \citep{diffprivacydwork2014algorithmic}. Machine unlearning is strongly intertwined with this goal, with \citet{ginart2019making} introducing a probabilistic definition of unlearning that requires the output distribution of a model that has unlearnt data to be similar to the output distribution of a model that was never trained on that data.

\textbf{Membership inference attacks (MIA).} Deep learning models generally perform better on their training data than unseen data. Membership attacks exploit this to determine if a specific set of data was used in the training process by comparing model output distributions for test and train data \citet{shokri2017membership, hu2022membership}. MIA is therefore a key measure of performance for unlearning methods.

\textbf{Unlearning in deep networks.} Due to the high cost of training for large models, and the need to be applied to existing models, we restrict our review to post hoc methods that do not require additional computations or data storage during the original training process (e.g., gradient vectors in \citet{mehta2022deep}, or a summation layer in \citep{cao2015towards}). We categorise post hoc deep neural network unlearning methods into retraining-based and retraining-free approaches, based on whether they require any traditional model training steps in the unlearning procedure. 

\textbf{Retraining-free} unlearning methods commonly utilise the Fisher information matrix. The FIM has long been used to approximate the sensitivity of a model's output to perturbations of its parameters from the second derivative of the loss (i.e. the Hessian), which can be interpreted as the importance of each parameter (as in \citet{kirkpatrick2017overcoming} where it is used to calculate an $L_{2}$ regularization term to prevent forgetting of previous tasks). In unlearning, the FIM has been used in ad hoc  \cite{guo2019certified}, post hoc \cite{golatkar2020eternal}, and zero-shot \cite{sekhari2021remember} approaches.  \citet{golatkar2020eternal} introduces Fisher Forgetting, a weight scrubbing method that induces forgetting by injecting noise into the parameters proportional to their relative importance to the forget set compared to the retain set. This is computationally very expensive and updates the whole model which causes significant degradation to the accuracy on the retain dataset, as shown in experiments of \citet{tarun2023fast}. SSD addresses these shortcomings, yielding significantly faster execution time and through a stringent parameter-selection step, retain set performance is much better protected. Other non-FIM-based methods include variational forgetting for regression and Gaussian processes \citep{nguyen2020variational}, neural tangent kernel forgetting (NTK) \citep{golatkar2020forgetting}, and mixed-linear models (MLM) \citep{golatkar2021mixed}. NTK and MLM rely on additional models that add further complexity and overhead. Selective Synaptic Dampening does not rely on any additional models.

\textbf{Retraining-based} unlearning methods are the current state-of-the-art in terms of performance. 
\citet{chundawat2023can} uses a student-teacher framework with a competent and incompetent teacher model to induce forgetting while preserving model performance on the retained data.
\citet{graves2021amnesiac} present two unlearning methods of which one is post hoc which we will refer to as amnesiac in this paper. Amnesiac relabels $\mathcal{D}_f$ with randomly selected incorrect labels and then retrains the network for a set number of epochs.
\citet{tarun2023fast} learns an error-maximising noise matrix for $\mathcal{D}_f$ that is then applied to the weights in the impair step before performing a repair step to recover model performance on $\mathcal{D}_{r}$.
\citet{chundawat2023zero} and \citet{tarun2023deep} address the related yet distinct challenges of zero-shot and deep regression unlearning, respectively. We restrict the scope of this work to the more mature unlearning area of classification tasks.

Our method contrasts existing works by possessing a combination of desirable properties: post hoc, fast, retrain-free, selective in the parameters to be manipulated, and not reliant on additional models.

\section{Preliminaries}
Let $\mathcal{D}=\{x_{i}, y_{i}\}^{N}_{i=1}$ be a dataset of training samples $x_{i}$, with corresponding class label $y_{i} \in \{1,...,K\}$. In an unlearning scenario, the objective is to forget the subset $\mathcal{D}_{f} \subset \mathcal{D}$, while preserving model performance on the remaining data $\mathcal{D}_{r} = \mathcal{D} \setminus \mathcal{D}_{f} $. We shall refer to these subsets as the forget set and retain set, respectively. $\mathcal{D}_f$ may comprise any subset of $\mathcal{D}$, and we show the performance of SSD on full class forgetting, where the forget set contains all samples with label $k$, subclass forgetting, where a subset of samples with label $k$ are forgotten, and random subset, where each datapoint is randomly sampled from $\mathcal{D}$, without replacement.
Let $\phi_{\theta}(\cdot): X \rightarrow Y$, where $X \in \mathbb{R}^{n}$ and $Y \in \mathbb{R}^{K}$, be a function parameterised by $\theta \in \mathbb{R}^{m}$ and trained on $\mathcal{D}$, where the $k$-the component of $\phi_{\theta}(x) $ is the probability that sample $x $ belongs to class $k$.

\section{Proposed Method}
The guiding intuition behind Selective Synaptic Dampening is that there likely exist parameters that are specifically important for $\mathcal{D}_{f}$ but not for $\mathcal{D}_{r}$. This intuition is further motivated by works such as \citet{feldman2020does} and \citet{stephenson2021geometry}. They show that deep neural networks memorize specific training examples and that parameters in later layers are highly specialized to specific features. Such parameters are likely extremely important for a small set of samples in the training data, but may not be generally important for the wider training set. Since $\mathcal{D}_{r}$ is typically large and filled with diverse samples, parameters which are similarly or more important for $\mathcal{D}_{r}$ compared to $\mathcal{D}_{f}$ likely correspond to highly generalized features, with little to no threat to differential privacy or the right to be forgotten. For example, recognising that there exists a person in an image is not necessarily a problem, but identifying \textit{who} that person is, is a significant problem.
\subsubsection{Hessian and the Fisher information matrix}
One way to identify important parameters is to use the FIM, as in \citet{kirkpatrick2017overcoming, golatkar2020eternal, guo2019certified, lecun1989optimal, hassibi1993optimal}. Given $\phi_{\theta}$, it can be assumed that the optimal parameters $\theta^*$ have been learnt, which minimises the loss over $\mathcal{D}$. The sensitivity of $\phi_{\theta}$ with respect to each parameter $\theta_{k}$ can be calculated via the second-order derivative of the loss near the minimum \cite{maltoni2019continuous}. This sensitivity can be interpreted as the importance of each parameter. Calculating the second derivative is expensive, however, the diagonal of the Fisher information matrix is equivalent to the second derivative of the loss \cite{pawitan2001all}, and critically can be computed using first-order derivatives. The FIM, and its first-order derivative property \cite{kay1993fundamentals, aich2021elastic}, is given in Eq. \ref{eq:fisher}.

\begin{equation*}
[]_{\mathcal{D}} = \mathbb{E}  \left[ -\frac{\delta^{2} \ ln \ p(\mathcal{D}|\theta)}{\delta \theta^{2}} \left. \right\vert_{\theta^{*}_{\mathcal{D}}} \right] 
\end{equation*}
\begin{equation}
[]_{\mathcal{D}} = \mathbb{E}  \left[ \left( \left( \frac{\delta \ ln \ p(\mathcal{D}|\theta)}{\delta \theta} \right) \left( \frac{\delta \ ln \ p(\mathcal{D}|\theta)}{\delta \theta} \right)^{T} \right)\left. \right\vert_{\theta^{*}_{\mathcal{D}}} \right]
\label{eq:fisher}
\end{equation}

\subsubsection{Selective Synaptic Dampening}
We begin by outlining a na\"ive forgetting approach using the FIM in Eq. \ref{eq:naive} 
\begin{equation}
    \theta_{i} = 
        \begin{cases}    
            0, & \text{if } []_{\mathcal{D}_{f,i}} > 0\\
            \theta_{i}, & \text{if } []_{\mathcal{D}_{f,i}} = 0   
    \end{cases}  \quad
    \forall i \in [0,|\theta|]
\label{eq:naive}
\end{equation}


where $[]_{\mathcal{D}_{f}, i}$ is the $i$-th element of the diagonal of the Fisher information matrix, calculated over the forget set $\mathcal{D}_f$. This represents a simple pruning algorithm, which identifies the location of all parameters that have non-zero importance values, and sets their value to zero, thereby removing their contribution to the model output. While this would lead to forgetting over $\mathcal{D}_f$, it would also lead to the catastrophic degradation of performance on $\mathcal{D}_{r}$, due to the large overlap in important parameters for both sets and the fact it is highly likely that $[]_{\mathcal{D}_f}$ is greater than zero for a majority of parameters. The challenge, then, lies in maintaining the forgetting abilities of such a pruning algorithm while simultaneously protecting parameters important to the retain set. To achieve this, we introduce two significant amendments to the pruning algorithm that lead to strong forgetting and retain-set performance while maintaining fast execution time. First, a stricter selection criterion is implemented, considering the parameter importance to the retain set in Eq. \ref{eq:select} 
\begin{equation}
    \theta_{i} = 
        \begin{cases}    
            0, & \text{if } []_{\mathcal{D}_{f,i}} > \alpha[]_{\mathcal{D}_r,i}\\
            \theta_{i}, & \text{if } []_{\mathcal{D}_{f,i}} \leq \alpha[]_{\mathcal{D}_r,i}  
    \end{cases}\quad
    \forall i \in [0,|\theta|]
\label{eq:select}
\end{equation}

where, the hyper-parameter $\alpha$ allows control of how protective the selection should be. The updated selection criteria now greatly reduces the number of parameters chosen, only selecting parameters that are more important for $\mathcal{D}_f$ than $\mathcal{D}_{r}$. This step facilitates the identification of parameters that are highly specialized towards samples in the forget set, with $\alpha$ dictating how specialized they must be to be pruned. While this step is critical, there remains a clear limitation with this approach, which is the binary nature of the update rule. 
A parameter that is slightly over the threshold is treated the same as a parameter that is vastly more important to the forget set. This lack of granularity limits the forgetting-performance trade-off and necessitates a large $\alpha$ to maintain performance on $\mathcal{D}_{r}$, however a large $\alpha$ then significantly reduces the ability to forget $\mathcal{D}_f$ due to an unreasonably high bar for being considered specialized. Therefore, the pruning step is replaced by a dampening step that applies a penalty to the magnitude of the parameter proportional to its relative importance of $\mathcal{D}_f$ compared to $\mathcal{D}$ in Eq. \ref{eq:dampen}
\begin{equation*}
    \beta = min(\frac{\lambda []_{\mathcal{D},i}}{[]_{\mathcal{D}_{f,i}}}, 1)
\end{equation*}
\begin{equation}    
    \theta_{i} = 
        \begin{cases}    
            \beta\theta_{i}, & \text{if } []_{\mathcal{D}_{f,i}} > \alpha[]_{\mathcal{D},i}\\
            \theta_{i}, & \text{if } []_{\mathcal{D}_{f,i}} \leq \alpha[]_{\mathcal{D},i}  
    \end{cases}\quad
    \forall i \in [0,|\theta|]\quad\quad
\label{eq:dampen}
\end{equation}
where $\lambda$ is a hyper-parameter to control the level of protection. This is the final SSD procedure. Intuitively, if $\lambda=1$ then $\beta < 1$ for all parameters that are specialized towards $\mathcal{D}_{f}$. Therefore, $\beta \rightarrow 0$ as a parameter becomes more specialized for $\mathcal{D}_{f}$. Since $\lambda$ scales this update, this dampening factor is given an upper bound of 1 to prevent large $\lambda$ values from causing parameters to grow. The dampening effect, combined with the selection criteria, creates a granular method to forgetting that will almost completely remove highly-specialized parameters, protect generalized parameters, and proportionally dampen the parameters in between, thereby finding an acceptable compromise to this multi-objective problem. $\lambda$ and $\alpha$ offer control over whether to prioritise forgetting or protecting, as well as performance adjustments for different models and unlearning tasks. Finally, we highlight that $[]_{\mathcal{D}_{r}}$ is substituted with  $[]_{\mathcal{D}}$ in the selection step. This is because $[]_{\mathcal{D}_{r}}$ must be recalculated for every new forget request. $[]_\mathcal{D}$ can be calculated at any point after training before unlearning and only needs to be computed once, allowing for the training set to be discarded and only $[]_\mathcal{D}$ stored. With this substitution, the selection criteria can be thought of as trying to prevent the dampening from moving the parameter set away from the original, optimal parameters $\theta^{*}_{\mathcal{D}}$. This is a trade-off to optimise for speed of execution, and the solution remains accurate as typically $|\mathcal{D}_{f}| << |\mathcal{D}|$ and therefore the values for $[]_\mathcal{D}$ and $[]_{\mathcal{D}_{r}}$ are near identical.  We hypothesise that this is a valid approach for repeated forget requests, as only a small fraction of parameters are updated for each forget request, and therefore it would take many forget requests for $[]_\mathcal{D}$ and $[]_{\mathcal{D}_{r}}$ to diverge substantially. In the event of such divergence, the only consequence would be the false protection of now-purged parameters, since dampening will only reduce a parameter's contribution to model output.

The experimental results are all calculated with $\mathcal{D}$ rather than $\mathcal{D}_{r}$, and demonstrate the efficacy of this approach.

\begin{algorithm}[tb]
\caption{Selective Synaptic Dampening}
\label{alg:algorithm}
\textbf{Input}: $\phi_{\theta}$, $\mathcal{D}$, $\mathcal{D}_f$; optional to skip 1.: $[]_{\mathcal{D}}$\\
\textbf{Parameter}: $\alpha, \lambda$\\
\textbf{Output}: $\phi_{\theta'}$
\begin{algorithmic}[1] 
\STATE Calculate and store $[]_{\mathcal{D}}$ once. Discard $\mathcal{D}$.
\STATE Calculate $[]_{\mathcal{D}_f}$
\FOR{i in range $|\theta|$}
\IF {$[]_{\mathcal{D}_{f}, i} > \alpha []_{\mathcal{D}, i}$}
\STATE  $\theta'_{i} = min(\frac{\lambda []_{\mathcal{D},i}}{[]_{\mathcal{D}_{f,i}}}\theta_{i}, \theta_{i}) $
\ENDIF
\ENDFOR
\STATE \textbf{return} $\phi_{\theta'}$
\end{algorithmic}
\end{algorithm}


\section{Experimental Setup}

\textbf{Datasets used.} We evaluate our method on image classification using CIFAR10, CIFAR20, and CIFAR100 \citep{CIFAR_krizhevsky2010convolutional}, in line with \citet{golatkar2020eternal, chundawat2023can}. We forget the same classes from these datasets as \cite{chundawat2023can}. \citet{golatkar2020eternal} also use the VGG-Face dataset \cite{parkhi2015deep}, however, this is no longer accessible so we substitute it with the PinsFaceRecognition dataset \cite{pinsds}, which consists of 17,534 faces of 105 celebrities collected from Pinterest.

\textbf{Models used.} Following \citet{chundawat2023can}, we use ResNet18 \citep{resnet_he2016deep} and Vision Transformer \citep{ViT_dosovitskiy2021image} for the learning and unlearning tasks. Experiments were performed on NVIDIA RTX4090 with Intel Xeon processors. Models are trained with early stopping using a multi-step learning rate scheduler beginning at $lr=0.1$ and the Adam optimiser \cite{kingma2014adam} with Python 3, PyTorch, and Ubuntu 20.04.6 LTS.\\
\textbf{Evaluation Measures.} Analogous to \citet{chundawat2023can}, we use the following: 1) \textit{Accuracy on the forget and retain set:} To validate forgetting while retaining overall model performance. Listed as $\mathcal{D}_f$ and $\mathcal{D}_{r}$ in results tables. 2) \textit{Membership inference attack:} To investigate if information about the forget sample is still present in the model. We use the logistic regression MIA implementation from \citet{chundawat2023can}. We also consider an additional metric 3) \textit{Execution time (seconds):} to evaluate the timeliness of methods (denoted $t$ in results). 

\textbf{Unlearning tasks:} We benchmark across three different unlearning scenarios: (i) Single-class forgetting \citep{chundawat2023zero}, (ii) sub-class forgetting \citep{Golatkar_2020_CVPR, golatkar2020forgetting}, and (iii) random observations forgetting \citep{golatkar2021mixed}.  In (i) we forget a superclass of CIFAR 20, as well as a class out of CIFAR100 and PinsFaceRecognition. In (ii) we forget a CIFAR 20 subclass of a superclass (e.g., rocket out of vehicles). CIFAR20 superclasses have CIFAR100 classes as subclasses. In (iii) we forget a random subset of 100 samples from CIFAR10.

\textbf{Baselines used.} We compare SSD to the Fisher Forgetting algorithm \cite{golatkar2020eternal}, however initial results show that not only does Fisher Forgetting perform worse than SSD, it also is exceptionally slow (between $50-250$ times slower than SSD). Similarly, \citet{nguyen2020variational, golatkar2020forgetting, golatkar2021mixed} fit additional compute-intensive models or perform computations exceeding Fisher in computational time (e.g., NTK ), thus rendering them prohibitively expensive to run over all experiments. Therefore, we focus our comparison on state-of-the-art methods, which are in the retraining-based category.
We compare SSD against the following methods: (a) \textit{Baseline:} The unaltered model trained on $\mathcal{D}_{r} \cup \mathcal{D}_f$ (b) \textit{Finetune:} Finetuning the baseline model on $\mathcal{D}_{r}$ for 5 epochs, (c) \textit{Retraining:} Retraining the model from scratch on $\mathcal{D}_{r}$, (d) \textit{Bad Teacher \citep{chundawat2023can}}, (e) \textit{Amnesiac \citep{graves2021amnesiac}}, (f) \textit{UNSIR \citep{tarun2023fast}}. (f) is not designed for random observations forgetting and therefore excluded from this task. 

\textbf{SSD parameters.} We found hyper-parameters using $50$ runs of the TPE search from Optuna \cite{akiba2019optuna}, for values $\alpha \in [0.1, 100])$ and $\lambda \in [0.1, 5]$. We only conducted this search for the Rocket and Veh2 classes. We use $\lambda$=1 and $\alpha$=10 for all ResNet18 CIFAR tasks. For PinsFaceRecognition, we use $\alpha$=50 and $\lambda$=0.1 due to the much greater similarity between classes. ViT also uses $\lambda$=1 on all CIFAR tasks. We change $\alpha$=10 to $\alpha$=5 for slightly improved performance on class and $\alpha$=25 on sub-class unlearning.

\begin{figure}[t!]
\includegraphics[width=0.45\textwidth]{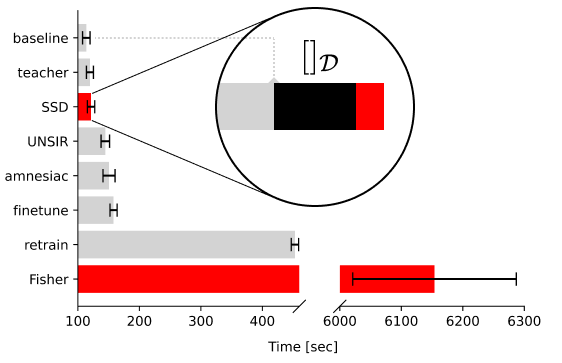}
\caption{Time per method on (ii) Cifar20 rocket including model loading and metric calculation (i.e. baseline) to simulate a realistic forget process from model loading to verification of forgetting. Zoom-in indicates share of compute time spent on  calculating $[]_{\mathcal{D}}$ compared to the rest of the SSD method. Precomputing $[]_{\mathcal{D}}$ can save 84.32\%$\pm$0.21\% of the pure SSD computing time (excl. baseline time).}
\label{fig:times}
\end{figure}

\begin{table}[t!]
\fontsize{9pt}{12pt}\selectfont
\setlength{\tabcolsep}{3.5pt}
\centering
\begin{tabular}{llccc}
\cline{3-5}
                                                                           &                     & retrain & Fisher & SSD \\ \hline
\multirow{4}{*}{\begin{tabular}[c]{@{}l@{}}Cifar20\\ Veh2\end{tabular}}    & $\mathcal{D}_{r}$ & \textbf{82.11$\pm$0.19}    & 5.76$\pm$1.01                & \textbf{82.97$\pm$0.00}   \\ 
                                                                           & $\mathcal{D}_f$             & \textbf{0.00$\pm$0.00}   & \textbf{0.00$\pm$0.00}               & \textbf{0.00$\pm$0.00}  \\ 
                                                                           & MIA          & \textbf{13.54$\pm$0.01}        & 47.12$\pm$16.39               & \textbf{6.68$\pm$0.00}   \\ 
                                                                           & $t$   &     441$\pm$10      &       5871$\pm$297           &   \textbf{122$\pm$5}  \\ \hline
\multirow{4}{*}{\begin{tabular}[c]{@{}l@{}}Cifar100\\ Rocket\end{tabular}} & $\mathcal{D}_{r}$ & \textbf{72.83$\pm$0.42}        & 1.18$\pm$0.06                & \textbf{74.54$\pm$0.00}   \\ 
                                                                           & $\mathcal{D}_f$            & \textbf{0.00$\pm$0.00}       & \textbf{0.00$\pm$0.00}               & \textbf{0.00$\pm$0.00}  \\ 
                                                                           & MIA           & \textbf{1.04$\pm$0.00}       & \textbf{0.00$\pm$0.16 }               & 2.20$\pm$0.00   \\ 
                                                                           & $t$                &   1805$\pm$10       &     28744$\pm$1332             &   \textbf{120$\pm$6}  \\ \hline
\multirow{4}{*}{\begin{tabular}[c]{@{}l@{}}Cifar20\\ Rocket\end{tabular}}  & $\mathcal{D}_{r}$ & \textbf{81.54$\pm$0.24}       & 5.20$\pm$0.54                & \textbf{82.43$\pm$0.00}   \\ 
                                                                           & $\mathcal{D}_f$             & \textbf{10.74$\pm$3.40}        & 0.78$\pm$1.35                & \textbf{2.17$\pm$0.00}   \\ 
                                                                           & MIA           & \textbf{3.85$\pm$0.01}        & 43.40$\pm$7.79                & \textbf{10.80$\pm$0.00}   \\ 
                                                                           & $t$                &       453$\pm$6   &       6154$\pm$133           &   \textbf{121$\pm$6}  \\ \hline
\multirow{4}{*}{\begin{tabular}[c]{@{}l@{}}Cifar10\\ Random\end{tabular}}  & $\mathcal{D}_{r}$ & \textbf{91.45$\pm$0.11}      & 12.74$\pm$2.22                & \textbf{88.68$\pm$3.36}  \\ 
                                                                           & $\mathcal{D}_f$            & \textbf{94.10$\pm$2.00}        & 11.85$\pm$4.13                & \textbf{93.61$\pm$4.99}   \\ 
                                                                           & MIA           & \textbf{74.22$\pm$0.04}        & 46.59$\pm$27.46                & \textbf{72.65$\pm$0.05}   \\ 
                                                                           & $t$                 &    308$\pm$5      &    3225$\pm$100              &  \textbf{121$\pm$5}   \\ \hline 
\end{tabular}
\caption{\textbf{Fisher} unlearning on \textbf{(i), (ii), and (iii) tasks} with \textbf{ResNet18}. As reported by \citet{tarun2023fast}, we also observe that Fisher fails to maintain accuracy on the retained data and is computationally very expensive. SSD is deterministic, thus no $\pm$ is reported. Veh2: Vehicle2. $\mathcal{D}_{r}$ and $\mathcal{D}_f$ rows report the accuracy on the respective dataset. All values in percent [\%] except $t$ [seconds]. For retrain, $D_f>0$ due to model generalization (e.g., rocket similar to vehicles)}
\label{tab:fisher_eternal}
\end{table}

\begin{table*}[t!]
\fontsize{9pt}{12pt}\selectfont
\centering
\begin{tabular}{ll|l|c|c|c|c|c|c|c}
\hline
                     & Class                   & metric & baseline       & retrain         & finetune        & teacher         & UNSIR           & amnesiac            & SSD              \\ \hline
\multirow{6}{*}{RN}  & \multirow{3}{*}{rocket} & $\mathcal{D}_{r}$  & \textbf{76.27$\pm$0.00} & 72.83$\pm$0.42 & 64.05$\pm$0.88 & 74.53$\pm$0.26 & 73.89$\pm$0.28 & 73.34$\pm$0.45 &\textbf{ 74.54$\pm$0.00}   \\
                     &                         & $\mathcal{D}_{f}$   & 80.90$\pm$0.00  & \textbf{0.00$\pm$0.00}    & 0.00$\pm$0.00    & \textbf{0.00$\pm$0.00}    & 28.66$\pm$4.98 & \textbf{0.00$\pm$0.00}    & \textbf{0.00$\pm$0.00  }   \\
                     &                         & MIA    & 93.40$\pm$0.00  & \textbf{1.04$\pm$0.41}   & 13.70$\pm$0.04  & \textbf{0.00$\pm$0.00}    & 1.94$\pm$0.01  & 29.56$\pm$0.02 & 2.20$\pm$0.00     \\ \cline{2-10} 
                     & \multirow{3}{*}{MR}     & $\mathcal{D}_{r}$  & \textbf{76.28$\pm$0.00} & 72.90$\pm$0.45  & 63.97$\pm$0.67 & 74.53$\pm$0.26 & 73.81$\pm$0.26 & 73.56$\pm$0.48 & \textbf{75.59$\pm$0.00 }  \\
                     &                         & $\mathcal{D}_{f}$   & 80.12$\pm$0.00 & \textbf{0.00$\pm$0.00 }   & 0.00$\pm$0.00    & \textbf{0.00$\pm$0.00}    & 27.34$\pm$5.08 & \textbf{0.00$\pm$0.00}    & \textbf{0.00$\pm$0.00 }    \\
                     &                         & MIA    & 95.20$\pm$0.00  & \textbf{0.22$\pm$0.01}   & 12.98$\pm$0.03 & 0.00$\pm$0.00    & 1.54$\pm$0.01  & 46.48$\pm$0.04 &\textbf{ 0.20$\pm$0.00  }   \\ \hline \hline
\multirow{6}{*}{ViT} & \multirow{3}{*}{rocket} & $\mathcal{D}_{r}$  & \textbf{88.88$\pm$0.00 }& 90.07$\pm$0.09 & 80.82$\pm$1.37 & 87.46$\pm$0.53 & 88.47$\pm$0.38 & 87.92$\pm$0.89 & \textbf{88.90$\pm$0.00}  \\
                     &                         & $\mathcal{D}_{f}$   & 94.70$\pm$0.00  & \textbf{0.00$\pm$0.00}    & 0.46$\pm$0.72  & 4.20$\pm$5.24   & 65.32$\pm$9.11 & \textbf{0.00$\pm$0.00 }   &\textbf{ 0.00$\pm$0.00}    \\
                     &                         & MIA    & 94.40$\pm$0.00  & \textbf{3.23$\pm$0.50}   & 19.00$\pm$0.09  & 0.03$\pm$0.00   & 29.13$\pm$0.06 & 1.00$\pm$0.01   & \textbf{1.80$\pm$0.00}    \\ \cline{2-10} 
                     & \multirow{3}{*}{MR}     & $\mathcal{D}_{r}$  & 88.87$\pm$0.00 & \textbf{90.02$\pm$0.22} & 81.14$\pm$0.79 & 87.42$\pm$0.41 & 88.44$\pm$0.58 & 88.34$\pm$0.72 & \textbf{88.82$\pm$0.00}  \\
                     &                         & $\mathcal{D}_{f}$   & 94.88$\pm$0.00 &\textbf{ 0.00$\pm$0.00 }   & 2.33$\pm$2.37  & 12.82$\pm$5.92 & 83.94$\pm$2.87 & \textbf{0.00$\pm$0.00}    & \textbf{0.00$\pm$0.00} \\
                     &                         & MIA    & 92.80$\pm$0.00  & \textbf{0.70$\pm$0.41}    & 7.10$\pm$0.02   & 0.03$\pm$0.00   & 21.33$\pm$0.03 & \textbf{0.47$\pm$0.00}   & 3.80$\pm$0.00   \\ \hline
\end{tabular}
\caption{\textbf{(i) Class} unlearning on \textbf{CIFAR100} with \textbf{ResNet18 (RN)} and \textbf{Vision Transformer (ViT)}. MR: mushroom. $\mathcal{D}_{r}$ and $\mathcal{D}_f$ rows report the accuracy on the respective dataset. All values in percent [\%].}
\label{tab:full_100}
\end{table*}

\begin{table*}[t!]
\fontsize{9pt}{12pt}\selectfont
\centering
\begin{tabular}{ll|l|c|c|c|c|c|c|c}
\hline
                     & Class                 & metric & baseline       & retrain         & finetune        & teacher         & UNSIR           & amnesiac            & SSD              \\ \hline
\multirow{6}{*}{RN}  & \multirow{3}{*}{Veh2} & $\mathcal{D}_{r}$  & \textbf{82.69$\pm$0.00} & 82.11$\pm$0.19 & 73.50$\pm$0.86  & 81.96$\pm$0.21 & 80.81$\pm$0.46 & 81.13$\pm$0.3  & \textbf{82.97$\pm$0.00}   \\
                     &                       & $\mathcal{D}_{f}$   & 80.41$\pm$0.00 & \textbf{0.00$\pm$0.00}    & 0.00$\pm$0.00    & 3.62$\pm$1.07  & 46.92$\pm$2.27 & \textbf{0.00$\pm$0.00}    & \textbf{0.00$\pm$0.00 }    \\
                     &                       & MIA    & 82.56$\pm$0.00 & \textbf{13.54$\pm$0.01} & 30.63$\pm$0.04 & 0.00$\pm$0.00    & 35.16$\pm$0.03 & \textbf{7.54$\pm$0.01}  & 6.68$\pm$0.00    \\ \cline{2-10} 
                     & \multirow{3}{*}{veg}  & $\mathcal{D}_{r}$  & \textbf{82.31$\pm$0.00} & 81.39$\pm$0.21 & 71.42$\pm$1.32 & 81.46$\pm$0.3  & 80.29$\pm$0.26 & 81.01$\pm$0.33 & \textbf{82.38$\pm$0.00}   \\
                     &                       & $\mathcal{D}_{f}$   & 86.90$\pm$0.00  & \textbf{0.00$\pm$0.00}    & 0.00$\pm$0.00    & 2.67$\pm$1.35  & 64.45$\pm$1.77 & \textbf{0.00$\pm$0.00}    & \textbf{0.00$\pm$0.00 }    \\
                     &                       & MIA    & 89.52$\pm$0.00 & \textbf{9.74$\pm$0.01}  & 29.39$\pm$0.08 & 0.00$\pm$0.00    & 40.66$\pm$0.06 & \textbf{5.00$\pm$0.01}   & 16.96$\pm$0.00   \\ \hline \hline
\multirow{6}{*}{ViT} & \multirow{3}{*}{Veh2} & $\mathcal{D}_{r}$  & \textbf{95.73$\pm$0.00} & 94.85$\pm$0.13 & 87.75$\pm$1.64 & 93.59$\pm$0.3  & 93.56$\pm$0.32 & \textbf{93.88$\pm$0.15} & 93.12$\pm$0.00   \\
                     &                       & $\mathcal{D}_{f}$   & 95.22$\pm$0.00 &\textbf{ 0.00$\pm$0.00}    & 0.04$\pm$0.12  & 4.88$\pm$4.12  & 70.31$\pm$5.03 & \textbf{0.00$\pm$0.00 }   & \textbf{0.00$\pm$0.00} \\
                     &                       & MIA    & 84.04$\pm$0.00 & \textbf{22.96$\pm$0.03} & 38.15$\pm$0.08 & 0.02$\pm$0.00   & 48.98$\pm$0.07 & 1.20$\pm$0.00    & \textbf{7.04$\pm$0.00 } \\ \cline{2-10} 
                     & \multirow{3}{*}{veg}  & $\mathcal{D}_{r}$  & \textbf{95.59$\pm$0.00} & 94.54$\pm$0.21 & 87.09$\pm$1.24 & 92.92$\pm$0.51 & 93.25$\pm$0.35 & 93.29$\pm$0.41 & \textbf{95.71$\pm$0.00}  \\
                     &                       & $\mathcal{D}_{f}$   & 97.57$\pm$0.00 & \textbf{0.00$\pm$0.00 }   & 0.30$\pm$0.29   & 8.28$\pm$6.79  & 89.02$\pm$2.41 & 0.02$\pm$0.07  & \textbf{0.00$\pm$0.00} \\
                     &                       & MIA    & 91.32$\pm$0.00 & \textbf{4.41$\pm$0.01}  & 14.72$\pm$0.05 & 0.02$\pm$0.00   & 58.67$\pm$0.04 & 1.02$\pm$0.00   & \textbf{1.88$\pm$0.00}  \\ \hline
\end{tabular}
\caption{\textbf{(i) Class} unlearning on \textbf{CIFAR20} with \textbf{ResNet18} and \textbf{Vision Transformer}. Veh2: Vehicle2.}
\label{tab:full_20}
\end{table*}

\begin{table*}[t!]
\fontsize{9pt}{12pt}\selectfont
\centering
\begin{tabular}{ll|c|c|c|c|c|c|c}
\hline
                    & metric & baseline        & retrain        & finetune        & teacher         & UNSIR           & amnesiac        & SSD     \\ \hline
\multirow{3}{*}{RN} & $\mathcal{D}_{r}$  & 98.52$\pm$0.02 & \textbf{100.00$\pm$0.00} & 99.72$\pm$0.45 & 96.72$\pm$0.44 & 99.89$\pm$0.06 & \textbf{99.99$\pm$0.02} & 98.42$\pm$0.13 \\
                    & $\mathcal{D}_{f}$   & 97.84$\pm$1.99 &\textbf{ 0.00$\pm$0.00}   & 4.32$\pm$4.61  & 0.13$\pm$0.4   & 90.53$\pm$5.68 &\textbf{ 0.00$\pm$0.00 }   & \textbf{0.00$\pm$0.00}    \\
                    & MIA    & 34.38$\pm$0.23 & \textbf{0.00$\pm$0.00}   & 0.80$\pm$0.01   &\textbf{ 0.02$\pm$0.00}   & 8.54$\pm$0.11  & 8.92$\pm$0.03  & 1.11$\pm$0.01  \\ \hline
\end{tabular}
\caption{\textbf{(i) Face} unlearning. One face unlearned per experiment [ID 1,10,20,30,40] and results aggregated for 5 experiments.}
\label{tab:face}
\end{table*}

\begin{table*}[t!]
\fontsize{9pt}{12pt}\selectfont
\centering
\begin{tabular}{ll|l|c|c|c|c|c|c|c}
\hline
                     & Class                   & metric & baseline       & retrain         & finetune        & teacher          & UNSIR            & amnesiac            & SSD            \\ \hline
\multirow{6}{*}{RN}  & \multirow{3}{*}{rocket} & $\mathcal{D}_{r}$  & \textbf{82.54$\pm$0.00} & 81.54$\pm$0.24 & 72.41$\pm$0.95 & 81.48$\pm$0.27  & 81.13$\pm$0.31  & 81.46$\pm$0.26 & \textbf{82.43$\pm$0.00} \\
                     &                         & $\mathcal{D}_{f}$   & 79.34$\pm$0.00 & \textbf{10.74$\pm$3.4}  & 9.75$\pm$6.68  & \textbf{6.41$\pm$3.57}   & 59.20$\pm$4.75   & 0.76$\pm$0.73  & 2.17$\pm$0.00  \\
                     &                         & MIA    & 89.40$\pm$0.00  & \textbf{3.85$\pm$0.01}  & 18.67$\pm$0.05 & 0.00$\pm$0.00     & 33.53$\pm$0.06  & \textbf{6.60$\pm$0.01}   & 10.80$\pm$0.00  \\ \cline{2-10} 
                     & \multirow{3}{*}{sea}    & $\mathcal{D}_{r}$  & \textbf{82.37$\pm$0.00} & 81.30$\pm$0.27  & 72.50$\pm$1.55  & 81.22$\pm$0.24  & 80.82$\pm$0.3   & 81.05$\pm$0.31 & \textbf{81.72$\pm$0.00} \\
                     &                         & $\mathcal{D}_{f}$   & 96.27$\pm$0.00 & \textbf{91.47$\pm$1.92} & 82.69$\pm$7.17 & 75.13$\pm$4.12  & \textbf{95.49$\pm$2.4}   & 46.78$\pm$8.55 & 75.35$\pm$0.00 \\
                     &                         & MIA    & 90.80$\pm$0.00  & \textbf{52.09$\pm$0.03} & 62.82$\pm$0.11 & 0.00$\pm$0.00     & \textbf{80.44$\pm$0.04}  & 4.45$\pm$0.01  & 21.80$\pm$0.00  \\ \hline \hline
\multirow{6}{*}{ViT} & \multirow{3}{*}{rocket} & $\mathcal{D}_{r}$  & \textbf{95.73$\pm$0.00} & 94.61$\pm$0.13 & 85.70$\pm$3.05  & 93.60$\pm$0.29   & 93.34$\pm$0.45  & 93.47$\pm$0.22 & \textbf{95.13$\pm$0.00} \\
                     &                         & $\mathcal{D}_{f}$   & 94.53$\pm$0.00 & \textbf{22.26$\pm$8.34} & 6.25$\pm$6.03  & 3.35$\pm$2.89   & 74.93$\pm$10.13 & 0.85$\pm$1.71  & \textbf{5.12$\pm$0.00} \\
                     &                         & MIA    & 80.40$\pm$0.00  & \textbf{3.44$\pm$0.01}  & 16.04$\pm$0.03 & 0.02$\pm$0.00    & 27.27$\pm$0.14  & \textbf{0.78$\pm$0.00}   &5.40$\pm$0.00   \\ \cline{2-10} 
                     & \multirow{3}{*}{sea}    & $\mathcal{D}_{r}$  & \textbf{95.67$\pm$0.00} & 94.55$\pm$0.22 & 87.65$\pm$1.56 & 93.57$\pm$0.26  & 93.26$\pm$0.31  & 93.26$\pm$0.24 & \textbf{95.57$\pm$0.00} \\
                     &                         & $\mathcal{D}_{f}$   & 99.22$\pm$0.00 & \textbf{95.12$\pm$0.81} & 89.17$\pm$4.17 & 25.97$\pm$14.01 & \textbf{94.25$\pm$2.32}  & 21.42$\pm$8.5  & 97.05$\pm$0.00 \\
                     &                         & MIA    & 88.40$\pm$0.00  & \textbf{65.96$\pm$0.04} & 65.04$\pm$0.13 & 0.17$\pm$0.00    & \textbf{76.96$\pm$0.07}  & 0.40$\pm$0.00    & 82.20$\pm$0.00  \\ \hline
\end{tabular}
\caption{\textbf{(ii) Subclass} unlearning on \textbf{CIFAR20} with \textbf{ResNet18} and \textbf{Vision Transformer}.}
\label{tab:sub_20}
\end{table*}

\begin{table*}[t!]
\fontsize{9pt}{12pt}\selectfont
\centering
\begin{tabular}{ll|l|c|c|c|c|c}
\hline
                     & metric & baseline        & retrain         & finetune        & teacher         & amnesiac            & \multicolumn{1}{l}{SSD} \\ \hline
\multirow{3}{*}{RN}  & $\mathcal{D}_{r}$  & 90.71$\pm$0.00  & \textbf{91.45$\pm$0.11} & 88.02$\pm$0.45 & \textbf{90.21$\pm$0.10}  & 90.16$\pm$0.23 & 88.68$\pm$3.36          \\
                     & $\mathcal{D}_{f}$   & 95.30$\pm$2.08  & \textbf{94.10$\pm$2.00}   & 90.00$\pm$3.73  & 90.00$\pm$2.73  & 59.04$\pm$4.79 & \textbf{93.61$\pm$4.99}          \\
                     & MIA    & 75.78$\pm$0.04 & \textbf{74.22$\pm$0.04} & 74.58$\pm$0.05 & 49.28$\pm$0.07 & 25.18$\pm$0.05 & \textbf{72.65$\pm$0.05}          \\ \hline \hline
\multirow{3}{*}{ViT} & $\mathcal{D}_{r}$  & \textbf{98.88$\pm$0.00}  & 98.61$\pm$0.08 & 97.28$\pm$0.33 & 97.58$\pm$0.36 & 97.62$\pm$0.35 & \textbf{98.01$\pm$1.56}          \\
                     & $\mathcal{D}_{f}$   & 100.00$\pm$0.00  & \textbf{98.80$\pm$0.76}  & 97.19$\pm$0.98 & 86.75$\pm$3.57 & 73.49$\pm$5.11 & \textbf{98.07$\pm$2.35}          \\
                     & MIA    & 90.76$\pm$0.03 & \textbf{91.77$\pm$0.02} & 86.14$\pm$0.02 & 33.53$\pm$0.06 & 10.44$\pm$0.05 & \textbf{85.54$\pm$0.11}          \\ \hline
\end{tabular}
\caption{\textbf{(iii) Random} unlearning on \textbf{CIFAR10} with \textbf{ResNet18} and \textbf{Vision Transformer.}}
\label{tab:random}
\end{table*}

\section{Results and Discussion}

\textbf{Defining good.} \citet{chundawat2023can} note that theoretically perfect accuracy $\mathcal{D}_f=0$ and $MIA=0$ is not necessarily ideal. Ideal is to closely match the performance of a model retrained from scratch that has never seen $\mathcal{D}_f$, the \textit{gold model}. They postulate that deviating from the gold model performance can lead to the \textit{Streisand effect}. They give the example of a model that classifies a Boeing aircraft as a mushroom. This is maximally wrong but not a behaviour expected from a model that can classify other aircraft. The model thus leaks information to an attacker by deliberately being wrong. The model has not unlearned, it simply learned to predict a wrong label for $\mathcal{D}_f$. This is congruent with the probabilistic definition of unlearning from \citet{ginart2019making}.
 This is especially relevant in the (iii) random forgetting scenario, where the distributions of $\mathcal{D}_{r}$ and $\mathcal{D}_f$ are likely to be very similar, and an unlearned sample from the rocket class could still be correctly classified based on the generalized knowledge about rockets in the model from the $\mathcal{D}_{r}$ rocket samples. Therefore, we define good as unlearning matching the MIA of the retrained model.

\textbf{Selectivity.} SSD only changes a small amount of parameters. When forgetting the rocket class from Cifar100, only $1.7\%$ of parameters are changed.

\textbf{Comparison to Fisher Forgetting.} Table \ref{tab:fisher_eternal} shows a comparison of Fisher \citep{golatkar2020eternal}, a retrained model, and SSD. Experimental results show that SSD significantly outperforms Fisher in terms of closeness to the retrained model performance, with Fisher significantly dropping $\mathcal{D}_{r}$ performance, and performing considerably worse on the MIA evaluation for the Cifar20 single-class and Cifar20 sublcass tasks. Furthermore, SSD is orders of magnitude faster than Fisher, only requiring $0.4 - 3.8$ percent of the time that Fisher requires. We also note that Fisher's execution time increases significantly with larger models, whereas SSD is less sensitive to this as shown with Cifar100 rocket class unlearning times of 120.00$\pm$5.49 seconds on ResNet8 (11M parameters) and 655.64$\pm$65.36 seconds on ViT (85M parameters).

 \textbf{Compute time comparison.} Fig. \ref{fig:times} shows compute times for the Cifar20 class unlearning task. We experimentally demonstrate that SSD is comparable to state-of-the-art methods. For repeated unlearning, as expected in practice, $[]_\mathcal{D}$ can be computed once and stored, reducing the time far below the already competitive time, which would make SSD the fastest method. Including the computation of $[]_\mathcal{D}$, SSD is the second fastest method behind \citet{chundawat2023can}. 

\textbf{(i) Class unlearning.} SSD is first benchmarked on class unlearning, as performed in \citet{chundawat2023can}, on CIFAR100 in Table \ref{tab:full_100}, CIFAR20 in Table \ref{tab:full_20}, and PinsFaceRecognition unlearning in Table \ref{tab:face}. SSD is close to the retrained model in terms of $\mathcal{D}_{r}$ and $MIA$ across the unlearning tasks and is comparable to retraining-based methods. For example, forgetting rocket from Cifar100 has a baseline MIA of $93\%$ for ResNet and $94\%$ for ViT, SSD reduces this to ca. $2\%$ (retrain $1$-$3\%$), while $\mathcal{D}_{r}$ performance drops just $2\%$ for ResNet, and actually improves negligibly for ViT. We highlight the closest method to retrain bold in the results tables.

\textbf{(ii) Subclass unlearning.} We present subclass unlearning, as performed in \citet{chundawat2023can}, on CIFAR20 in Table \ref{tab:sub_20}. Class \textit{sea} demonstrates the problem of defining good, as the retrained model achieves a high non-zero MIA. Amnesiac and Bad Teacher reduce MIA to near zero, even though a retrained model does not show the same behaviour. \citet{graves2021amnesiac} relabel $\mathcal{D}_{f}$ to random labels and \cite{chundawat2023can} uses an incompetent teacher to update the model. The noise-based approach of \citet{tarun2023fast} and our SSD on the other hand lead to higher MIA and $\mathcal{D}_f$ values that are closer to the retrained model. Efficacy analysis is therefore hard, as while Amnesiac and Teacher minimise the MIA and $\mathcal{D}_{f}$ accuracy, they may be falling victim to the \textit{Streisand effect}.

\textbf{(iii) Random sample unlearning.} We present random sample unlearning, as performed in \citet{golatkar2020eternal}, on CIFAR10 in Table \ref{tab:sub_20}. We observe similar performance of SSD and retraining on ResNet and ViT. \citet{graves2021amnesiac} and \citet{chundawat2023can} again reduce the MIA far below the retrained model as observed in (ii), which is a risk to privacy (\textit{Streisand effect}). An ideal unlearned model mimics a retrained model.

\textbf{Overall analysis of Selective Synaptic Dampening.} SSD outperformed Fisher Forgetting while being orders of magnitude faster, highlighting its efficacy. SSD also performs competitively with established state-of-the-art methods and full model retraining, demonstrating the viability of retrain-free post-hoc unlearning approaches in a wider context. SSD was, on average, the strongest performing method when measuring similarity to the fully retrained model. However, the lack of standardized evaluations in unlearning, and an as-yet-undecided notion of what is truly a \textit{good} MIA score, renders a qualitative assessment of methods challenging and determining the best algorithm ambiguous.

\textbf{Limitations.} First, SSD is not certified, with no mathematical guarantee of unlearning a given sample. This weakness is shared by all benchmarked methods. Second, if bad parameter values are chosen ($\alpha$, $\lambda$), such that large changes are made to the model, then repeat forgetting may lead to significant model degradation. Finding appropriate values for $\alpha$ and $\lambda$ is a practical limitation but as shown experimentally, the parameters are only set within one order of magnitude ($\alpha \in [5, 50]$ and $\lambda \in [0.1, 1]$) across two models of vastly different parameter counts and architectures. We hypothesise that the ideal parameters could be estimated from the $\mathcal{D}_f$ loss distribution to enable automatic parameter selection in future work. Finally, we note that without a repair step, there is naturally a finite amount of forget requests that SSD can process before $\mathcal{D}_{r}$ performance begins to degrade.

\section{Conclusion}


We present a novel two-step, retraining-free unlearning method. SSD first selects parameters that are considerably more important for the forget set than the retain set, before dampening these parameters proportional to the discrepancy in their importance to the forget and retain set. The result of these steps is a fast yet highly effective method for machine unlearning. We evaluate SSD on a range of tasks, demonstrating viability in single-class, sub-class and random sample settings, on multiple datasets and different model architectures. Results show that SSD is orders of magnitude faster than the comparable Fisher Forgetting method, outperforming the method considerably; SSD even rivals the speed and performance of state-of-the-art retrain-based approaches. 

Many future directions could be explored, 
such as evaluating how to improve and measure performance on random subsets, given the significant overlap in parameter importance for the forget and test set. Another interesting direction is how to forget large subsets of information. Typically experiments evaluate forgetting no more than 5-10\% of data; this may be realistic but evaluating how to increase the upper bound of forgetting without retraining may offer valuable insight into how to improve existing unlearning methods.

\section{Acknowledgments}

This work was supported by the Accenture Turing Strategic Partnership, the Turing Enrichment scheme, EPSRC CDT AgriFoRwArdS [grant number EP/S023917/1], and EPSRC DTP [grant number EP/W524633/1].

\bibliography{aaai24}

\appendix

\section{Appendix}

A single backward pass over the full dataset $D$ is $O(b_{full} \cdot p)$ where $b_{full}$ is the number of batches in the dataset and $p$ is the number of parameters in the model. Similarly, the complexity for the forget and retain set is $O(b_{forget} \cdot p)$ and $O(b_{retain} \cdot p)$, respectively. Note that $b_{retain} \approx b_{full} >> b_{forget}$.    
SSD requires only one computation of the gradients for $D$ (can be saved and reused), then for every forget request it requires one computation for $D_f$ to then apply the dampening in $O(b_{forget} \cdot p)$. Other methods are retraining-based and depend on the number of epochs $N$, generally on the order of $O(N \cdot b_{retain} \cdot p)$, which is much larger. 
\begin{table*}[t!]
\fontsize{9pt}{12pt}\selectfont
\setlength{\tabcolsep}{1.3pt}
\centering
\begin{tabular}{ll|l|c|c|c|c|c|c|c}
\hline
                      & Class                   & metric & baseline           & retrain             & finetune            & teacher          & UNSIR               & amnesiac            & SSD               \\ \hline
\multirow{20}{*}{RN}  & \multirow{4}{*}{baby}   & $\mathcal{D}_{r}$  & 82.45$\pm$0.00    & 81.45$\pm$0.3      & 72.94$\pm$0.82     & 81.25$\pm$0.19    & 80.91$\pm$0.31     & 81.02$\pm$0.2      & 79.25$\pm$0.00   \\
                      &                         & $\mathcal{D}_{f}$   & 86.98$\pm$0.00    & 80.03$\pm$3.4      & 66.52$\pm$6.19     & 75.17$\pm$2.66    & 82.73$\pm$3.27     & 37.70$\pm$9.59      & 8.85$\pm$0.00    \\
                      &                         & MIA    & 90.40$\pm$0.00     & 44.82$\pm$0.02     & 56.20$\pm$0.06      & 0.04$\pm$0.00     & 75.76$\pm$0.03     & 13.65$\pm$0.02     & 1.40$\pm$0.00     \\
                      &                         & $t$    & 113.43$\pm$4.57   & 449.69$\pm$3.52    & 158.26$\pm$6.14    & 120.18$\pm$4.92   & 141.03$\pm$5.82    & 147.22$\pm$8.63    & 118.88$\pm$4.87  \\ \cline{2-10} 
                      & \multirow{4}{*}{lamp}   & $\mathcal{D}_{r}$  & 82.61$\pm$0.00    & 81.81$\pm$0.19     & 72.46$\pm$0.95     & 81.51$\pm$0.19    & 81.12$\pm$0.27     & 81.36$\pm$0.36     & 79.58$\pm$0.00   \\
                      &                         & $\mathcal{D}_{f}$   & 72.66$\pm$0.00    & 22.95$\pm$4.72     & 17.25$\pm$6.58     & 16.82$\pm$3.07    & 50.44$\pm$5.76     & 7.44$\pm$1.73      & 0.00$\pm$0.00     \\
                      &                         & MIA    & 83.40$\pm$0.00     & 4.29$\pm$0.01      & 21.55$\pm$0.04     & 0.04$\pm$0.00     & 38.64$\pm$0.06     & 15.73$\pm$0.03     & 2.00$\pm$0.00     \\
                      &                         & $t$    & 111.93$\pm$6.1    & 452.52$\pm$7.55    & 154.76$\pm$3.83    & 119.23$\pm$6.96   & 142.33$\pm$6.24    & 144.30$\pm$5.62     & 120.68$\pm$6.13  \\ \cline{2-10} 
                      & \multirow{4}{*}{MR}     & $\mathcal{D}_{r}$  & 82.60$\pm$0.00     & 81.66$\pm$0.21     & 72.51$\pm$1.15     & 81.50$\pm$0.2      & 81.07$\pm$0.29     & 81.44$\pm$0.18     & 80.05$\pm$0.00   \\
                      &                         & $\mathcal{D}_{f}$   & 72.22$\pm$0.00    & 10.48$\pm$2.02     & 7.46$\pm$4.89      & 8.77$\pm$2.79     & 50.56$\pm$4.18     & 1.09$\pm$1.13      & 0.78$\pm$0.00    \\
                      &                         & MIA    & 86.80$\pm$0.00     & 2.27$\pm$0.00      & 18.09$\pm$0.02     & 0.00$\pm$0.00      & 27.29$\pm$0.05     & 12.80$\pm$0.03      & 4.00$\pm$0.00     \\
                      &                         & $t$    & 111.12$\pm$4.08   & 453.35$\pm$7.26    & 158.03$\pm$5.82    & 119.99$\pm$4.36   & 138.77$\pm$5.99    & 147.35$\pm$8.71    & 119.47$\pm$4.36  \\ \cline{2-10} 
                      & \multirow{4}{*}{Rkt} & $\mathcal{D}_{r}$  & 82.54$\pm$0.00    & 81.54$\pm$0.24     & 72.41$\pm$0.95     & 81.48$\pm$0.27    & 81.13$\pm$0.31     & 81.46$\pm$0.26     & 82.43$\pm$0.00   \\
                      &                         & $\mathcal{D}_{f}$   & 79.34$\pm$0.00    & 10.74$\pm$3.4      & 9.75$\pm$6.68      & 6.41$\pm$3.57     & 59.20$\pm$4.75      & 0.76$\pm$0.73      & 2.17$\pm$0.00    \\
                      &                         & MIA    & 89.40$\pm$0.00     & 3.85$\pm$0.01      & 18.67$\pm$0.05     & 0.00$\pm$0.00      & 33.53$\pm$0.06     & 6.60$\pm$0.01       & 10.80$\pm$0.00    \\
                      &                         & $t$    & 113.74$\pm$6.14   & 453.31$\pm$6.04    & 157.53$\pm$5.75    & 119.63$\pm$5.66   & 144.59$\pm$6.92    & 150.63$\pm$9.76    & 121.36$\pm$5.99  \\ \cline{2-10} 
                      & \multirow{4}{*}{sea}    & $\mathcal{D}_{r}$  & 82.37$\pm$0.00    & 81.30$\pm$0.27      & 72.50$\pm$1.55      & 81.22$\pm$0.24    & 80.82$\pm$0.3      & 81.05$\pm$0.31     & 81.72$\pm$0.00   \\
                      &                         & $\mathcal{D}_{f}$   & 96.27$\pm$0.00    & 91.47$\pm$1.92     & 82.69$\pm$7.17     & 75.13$\pm$4.12    & 95.49$\pm$2.4      & 46.78$\pm$8.55     & 75.35$\pm$0.00   \\
                      &                         & MIA    & 90.80$\pm$0.00     & 52.09$\pm$0.03     & 62.82$\pm$0.11     & 0.00$\pm$0.00      & 80.44$\pm$0.04     & 4.45$\pm$0.01      & 21.80$\pm$0.00    \\
                      &                         & $t$    & 110.59$\pm$4.79   & 457.37$\pm$8.39    & 156.36$\pm$5.69    & 122.91$\pm$7.55   & 140.82$\pm$6.64    & 149.29$\pm$9.85    & 118.85$\pm$4.42  \\ \hline \hline
\multirow{20}{*}{ViT} & \multirow{4}{*}{baby}   & $\mathcal{D}_{r}$  & 95.69$\pm$0.00    & 94.50$\pm$0.19      & 87.60$\pm$0.75      & 92.98$\pm$0.53    & 93.18$\pm$0.32     & 93.35$\pm$0.26     & 95.54$\pm$0.00           \\
                      &                         & $\mathcal{D}_{f}$   & 96.44$\pm$0.00    & 93.23$\pm$1.09     & 85.45$\pm$4.47     & 46.66$\pm$17.88   & 94.50$\pm$0.84      & 38.76$\pm$7.36     & 94.10$\pm$0.00              \\
                      &                         & MIA    & 91.60$\pm$0.00     & 77.37$\pm$0.03     & 66.57$\pm$0.07     & 0.03$\pm$0.00     & 88.00$\pm$0.03      & 0.93$\pm$0.01      & 77.20$\pm$0.00              \\
                      &                         & $t$    & 443.01$\pm$48.73  & 2148.68$\pm$112.52 & 1426.91$\pm$72.67  & 553.90$\pm$65.85   & 871.07$\pm$60.69   & 1024.99$\pm$35.22  & 736.94$\pm$12.01 \\ \cline{2-10} 
                      & \multirow{4}{*}{lamp}   & $\mathcal{D}_{r}$  & 95.77$\pm$0.00    & 94.69$\pm$0.13     & 87.72$\pm$0.52     & 93.57$\pm$0.67    & 93.36$\pm$0.53     & 93.66$\pm$0.5      & 95.54$\pm$0.00             \\
                      &                         & $\mathcal{D}_{f}$   & 89.58$\pm$0.00    & 34.55$\pm$8.62     & 16.90$\pm$10.4      & 8.23$\pm$7.05     & 76.48$\pm$5.18     & 0.59$\pm$1.45      & 14.58$\pm$0.00             \\
                      &                         & MIA    & 81.00$\pm$0.00     & 5.60$\pm$0.02       & 14.73$\pm$0.04     & 0.10$\pm$0.00      & 36.47$\pm$0.12     & 2.00$\pm$0.01       & 3.2$\pm$0.00               \\
                      &                         & $t$    & 419.69$\pm$25.19  & 2149.33$\pm$124.4  & 1424.87$\pm$71.39  & 558.10$\pm$63.92   & 899.77$\pm$72.01   & 1009.00$\pm$37.7    & 728.98$\pm$73.07 \\ \cline{2-10} 
                      & \multirow{4}{*}{MR}     & $\mathcal{D}_{r}$  & 95.69$\pm$0.00    & 94.60$\pm$0.13      & 87.37$\pm$0.91     & 93.55$\pm$0.42    & 93.15$\pm$0.6      & 93.42$\pm$0.46     & 95.51$\pm$0.00             \\
                      &                         & $\mathcal{D}_{f}$   & 97.05$\pm$0.00    & 26.57$\pm$6.41     & 15.71$\pm$12.13    & 13.01$\pm$9.11    & 79.77$\pm$7.58     & 0.20$\pm$0.36       & 6.68$\pm$0.00              \\
                      &                         & MIA    & 77.80$\pm$0.00     & 2.34$\pm$0.01      & 9.25$\pm$0.04      & 0.00$\pm$0.00      & 19.00$\pm$0.07      & 1.50$\pm$0.01       & 0.40$\pm$0.00               \\
                      &                         & $t$    & 525.69$\pm$124.31 & 2161.50$\pm$81.61   & 1442.36$\pm$113.19 & 620.06$\pm$111.74 & 925.70$\pm$117.5    & 1131.76$\pm$137.61 & 718.83$\pm$73.41 \\ \cline{2-10} 
                      & \multirow{4}{*}{Rkt} & $\mathcal{D}_{r}$  & 95.73$\pm$0.00    & 94.61$\pm$0.13     & 85.70$\pm$3.05      & 93.60$\pm$0.29     & 93.34$\pm$0.45     & 93.47$\pm$0.22     & 95.13$\pm$0.00             \\
                      &                         & $\mathcal{D}_{f}$   & 94.53$\pm$0.00    & 22.26$\pm$8.34     & 6.25$\pm$6.03      & 3.35$\pm$2.89     & 74.93$\pm$10.13    & 0.85$\pm$1.71      & 5.12$\pm$0.00              \\
                      &                         & MIA    & 80.40$\pm$0.00     & 3.44$\pm$0.01      & 16.04$\pm$0.03     & 0.02$\pm$0.00     & 27.27$\pm$0.14     & 0.78$\pm$0.00      & 5.40$\pm$0.00               \\
                      &                         & $t$    & 535.00$\pm$75.3    & 2168.27$\pm$118.38 & 1436.63$\pm$117.86 & 631.88$\pm$115.25 & 983.06$\pm$143.25  & 1186.64$\pm$107.47 & 699.33$\pm$72.47 \\ \cline{2-10} 
                      & \multirow{4}{*}{sea}    & $\mathcal{D}_{r}$  & 95.67$\pm$0.00    & 94.55$\pm$0.22     & 87.65$\pm$1.56     & 93.57$\pm$0.26    & 93.26$\pm$0.31     & 93.26$\pm$0.24     & 95.57$\pm$0.00             \\
                      &                         & $\mathcal{D}_{f}$   & 99.22$\pm$0.00    & 95.12$\pm$0.81     & 89.17$\pm$4.17     & 25.97$\pm$14.01   & 94.25$\pm$2.32     & 21.42$\pm$8.5      & 97.05$\pm$0.00             \\
                      &                         & MIA    & 88.40$\pm$0.00     & 65.96$\pm$0.04     & 65.04$\pm$0.13     & 0.17$\pm$0.00     & 76.96$\pm$0.07     & 0.40$\pm$0.00       & 82.20$\pm$0.00              \\
                      &                         & $t$    & 475.46$\pm$120.27 & 2142.05$\pm$85.86  & 1512.88$\pm$129.82 & 586.67$\pm$89.72  & 1024.45$\pm$144.84 & 1070.37$\pm$138.06 & 645.74$\pm$53.38 \\ \hline
\end{tabular}
\caption{Cifar20 subclass unlearning. Parameters: ResNet18 $\alpha = 10, \lambda = 1$, Vision Transformer $\alpha = 25, \lambda = 1$. MR: mushrooms, Rkt: rocket.; $\mathcal{D}_{r}$ and $\mathcal{D}_f$ report the accuracy on the respective dataset in percent. MIA in percent and time $t$ in seconds.}
\label{tab:appendix}
\end{table*}

\begin{table*}[t!]
\fontsize{9pt}{12pt}\selectfont
\setlength{\tabcolsep}{1.3pt}
\centering
\begin{tabular}{ll|l|c|c|c|c|c|c|c}
\hline
                      & Class                   & metric & baseline          & retrain             & finetune            & teacher          & UNSIR               & amnesiac            & SSD                \\ \hline
\multirow{20}{*}{RN}  & \multirow{4}{*}{baby}   & $\mathcal{D}_{r}$  & 76.38$\pm$0.00   & 73.10$\pm$0.55      & 64.03$\pm$0.8      & 74.71$\pm$0.19    & 74.16$\pm$0.29     & 73.69$\pm$0.27     & 44.05*$\pm$0.00    \\
                      &                         & $\mathcal{D}_{f}$   & 72.48$\pm$0.00   & 0.00$\pm$0.00       & 0.00$\pm$0.00       & 0.08$\pm$0.25     & 11.70$\pm$4.52      & 0.00$\pm$0.00       & 0.00$\pm$0.00      \\
                      &                         & MIA    & 92.60$\pm$0.00    & 2.44$\pm$0.01      & 24.52$\pm$0.07     & 0.00$\pm$0.00      & 4.88$\pm$0.03      & 50.86$\pm$0.05     & 5.40$\pm$0.00      \\
                      &                         & $t$    & 111.62$\pm$4.7   & 1808.49$\pm$12.3   & 160.27$\pm$6.16    & 118.27$\pm$4.83   & 138.78$\pm$4.02    & 150.12$\pm$8.79    & 121.02$\pm$5.19   \\ \cline{2-10} 
                      & \multirow{4}{*}{lamp}   & $\mathcal{D}_{r}$  & 76.39$\pm$0.00   & 72.89$\pm$0.34     & 64.01$\pm$0.63     & 74.76$\pm$0.19    & 73.93$\pm$0.26     & 73.52$\pm$0.48     & 76.08$\pm$0.00    \\
                      &                         & $\mathcal{D}_{f}$   & 70.49$\pm$0.00   & 0.00$\pm$0.00       & 0.00$\pm$0.00       & 0.00$\pm$0.00      & 15.04$\pm$2.58     & 0.00$\pm$0.00       & 0.00$\pm$0.00      \\
                      &                         & MIA    & 92.40$\pm$0.00    & 0.32$\pm$0.00      & 12.92$\pm$0.04     & 0.00$\pm$0.00      & 0.96$\pm$0.01      & 46.24$\pm$0.04     & 0.20$\pm$0.00      \\
                      &                         & $t$    & 110.75$\pm$3.52  & 1807.21$\pm$6.58   & 156.66$\pm$5.56    & 117.02$\pm$2.7    & 141.85$\pm$4.67    & 145.52$\pm$5.49    & 120.56$\pm$5.92   \\ \cline{2-10} 
                      & \multirow{4}{*}{MR}     & $\mathcal{D}_{r}$  & 76.28$\pm$0.00   & 72.90$\pm$0.45      & 63.97$\pm$0.67     & 74.53$\pm$0.26    & 73.81$\pm$0.26     & 73.56$\pm$0.48     & 75.59$\pm$0.00    \\
                      &                         & $\mathcal{D}_{f}$   & 80.12$\pm$0.00   & 0.00$\pm$0.00       & 0.00$\pm$0.00       & 0.00$\pm$0.00      & 27.34$\pm$5.08     & 0.00$\pm$0.00       & 0.00$\pm$0.00      \\
                      &                         & MIA    & 95.20$\pm$0.00    & 0.22$\pm$0.00      & 12.98$\pm$0.03     & 0.00$\pm$0.00      & 1.54$\pm$0.01      & 46.48$\pm$0.04     & 0.20$\pm$0.00      \\
                      &                         & $t$    & 110.58$\pm$2.51  & 1810.03$\pm$12.62  & 156.15$\pm$5.83    & 119.11$\pm$5.09   & 137.77$\pm$5.32    & 146.45$\pm$7.43    & 118.33$\pm$4.26   \\ \cline{2-10} 
                      & \multirow{4}{*}{Rkt} & $\mathcal{D}_{r}$  & 76.27$\pm$0.00   & 72.83$\pm$0.42     & 64.05$\pm$0.88     & 74.53$\pm$0.26    & 73.89$\pm$0.28     & 73.34$\pm$0.45     & 74.54$\pm$0.00    \\
                      &                         & $\mathcal{D}_{f}$   & 80.90$\pm$0.00    & 0.00$\pm$0.00       & 0.00$\pm$0.00       & 0.00$\pm$0.00      & 28.66$\pm$4.98     & 0.00$\pm$0.00       & 0.00$\pm$0.00      \\
                      &                         & MIA    & 93.40$\pm$0.00    & 1.04$\pm$0.00      & 13.70$\pm$0.04      & 0.00$\pm$0.00      & 1.94$\pm$0.01      & 29.56$\pm$0.02     & 2.20$\pm$0.00      \\
                      &                         & $t$    & 111.23$\pm$4.66  & 1804.93$\pm$9.76   & 159.25$\pm$3.84    & 120.36$\pm$6.71   & 138.25$\pm$5.74    & 146.64$\pm$7.62    & 120.00$\pm$5.49    \\ \cline{2-10} 
                      & \multirow{4}{*}{sea}    & $\mathcal{D}_{r}$  & 76.23$\pm$0.00   & 72.83$\pm$0.54     & 63.80$\pm$1.36      & 74.56$\pm$0.18    & 73.71$\pm$0.25     & 73.14$\pm$0.42     & 73.56$\pm$0.00    \\
                      &                         & $\mathcal{D}_{f}$   & 85.85$\pm$0.00   & 0.00$\pm$0.00       & 0.00$\pm$0.00       & 0.08$\pm$0.25     & 24.10$\pm$8.09      & 0.00$\pm$0.00       & 0.00$\pm$0.00      \\
                      &                         & MIA    & 93.40$\pm$0.00    & 5.84$\pm$0.02      & 26.54$\pm$0.08     & 0.02$\pm$0.00     & 4.80$\pm$0.02       & 29.50$\pm$0.03      & 0.60$\pm$0.00      \\
                      &                         & $t$    & 110.47$\pm$4.86  & 1843.27$\pm$91.85  & 156.51$\pm$5.35    & 118.19$\pm$4.88   & 162.51$\pm$77.78   & 143.35$\pm$4.72    & 119.77$\pm$4.75   \\ \hline \hline
\multirow{20}{*}{ViT} & \multirow{4}{*}{baby}   & $\mathcal{D}_{r}$  & 88.93$\pm$0.00   & 90.27$\pm$0.15     & 80.74$\pm$1.38     & 87.48$\pm$0.41    & 88.78$\pm$0.38     & 88.43$\pm$0.71     & 88.59$\pm$0.00              \\
                      &                         & $\mathcal{D}_{f}$   & 90.19$\pm$0.00   & 0.00$\pm$0.00       & 0.00$\pm$0.00       & 23.80$\pm$22.51    & 1.97$\pm$1.23      & 0.00$\pm$0.00       & 0.00$\pm$0.00                  \\
                      &                         & MIA    & 75.60$\pm$0.00    & 21.53$\pm$0.03     & 26.77$\pm$0.13     & 0.00$\pm$0.00      & 14.30$\pm$0.06      & 1.83$\pm$0.00      & 0.60$\pm$0.00                \\
                      &                         & $t$    & 459.67$\pm$83.01 & 2029.08$\pm$169.67 & 1461.82$\pm$96.24  & 610.04$\pm$98.05  & 954.61$\pm$110.27  & 1131.98$\pm$147.14 & 685.40$\pm$97.35   \\ \cline{2-10} 
                      & \multirow{4}{*}{lamp}   & $\mathcal{D}_{r}$  & 88.84$\pm$0.00   & 90.10$\pm$0.19      & 80.25$\pm$1.48     & 87.50$\pm$0.43     & 88.51$\pm$0.41     & 88.43$\pm$0.6      & 89.06$\pm$0.00              \\
                      &                         & $\mathcal{D}_{f}$   & 97.22$\pm$0.00   & 0.00$\pm$0.00       & 0.36$\pm$0.89      & 25.25$\pm$12.5    & 70.86$\pm$4.45     & 0.00$\pm$0.00       & 36.89$\pm$0.00              \\
                      &                         & MIA    & 95.60$\pm$0.00    & 2.27$\pm$0.01      & 11.77$\pm$0.04     & 0.13$\pm$0.00     & 29.40$\pm$0.05      & 2.70$\pm$0.00       & 0.40$\pm$0.00                \\
                      &                         & $t$    & 504.10$\pm$131.67 & 2115.02$\pm$196.32 & 1537.35$\pm$133.25 & 668.55$\pm$135.1  & 1002.18$\pm$135.09 & 1146.51$\pm$97.54  & 741.23$\pm$170.57 \\ \cline{2-10} 
                      & \multirow{4}{*}{MR}     & $\mathcal{D}_{r}$  & 88.87$\pm$0.00   & 90.02$\pm$0.22     & 81.14$\pm$0.79     & 87.42$\pm$0.41    & 88.44$\pm$0.58     & 88.34$\pm$0.72     & 88.82$\pm$0.00              \\
                      &                         & $\mathcal{D}_{f}$   & 94.88$\pm$0.00   & 0.00$\pm$0.00       & 2.33$\pm$2.37      & 12.82$\pm$5.92    & 83.94$\pm$2.87     & 0.00$\pm$0.00       & 0.00$\pm$0.00                  \\
                      &                         & MIA    & 92.80$\pm$0.00    & 0.70$\pm$0.00       & 7.10$\pm$0.02       & 0.03$\pm$0.00     & 21.33$\pm$0.03     & 0.47$\pm$0.00      & 3.80$\pm$0.00                \\
                      &                         & $t$    & 474.63$\pm$60.2  & 1948.82$\pm$123.1  & 1421.67$\pm$80.54  & 602.92$\pm$33.17  & 891.80$\pm$69.86    & 1071.85$\pm$98.15  & 650.93$\pm$76.43  \\ \cline{2-10} 
                      & \multirow{4}{*}{Rkt} & $\mathcal{D}_{r}$  & 88.88$\pm$0.00   & 90.07$\pm$0.09     & 80.82$\pm$1.37     & 87.46$\pm$0.53    & 88.47$\pm$0.38     & 87.92$\pm$0.89     & 88.90$\pm$0.00               \\
                      &                         & $\mathcal{D}_{f}$   & 94.70$\pm$0.00    & 0.00$\pm$0.00       & 0.46$\pm$0.72      & 4.20$\pm$5.24      & 65.32$\pm$9.11     & 0.00$\pm$0.00       & 0.00$\pm$0.00                  \\
                      &                         & MIA    & 94.40$\pm$0.00    & 3.23$\pm$0.00      & 19.00$\pm$0.09      & 0.03$\pm$0.00     & 29.13$\pm$0.06     & 1.00$\pm$0.01       & 1.80$\pm$0.00                \\
                      &                         & $t$    & 470.33$\pm$85.71 & 1939.74$\pm$109.65 & 1404.92$\pm$82.76  & 602.64$\pm$63.23  & 868.82$\pm$47.88   & 1029.67$\pm$70.8   & 655.64$\pm$65.36  \\ \cline{2-10} 
                      & \multirow{4}{*}{sea}    & $\mathcal{D}_{r}$  & 88.91$\pm$0.00   & 90.27$\pm$0.21     & 80.82$\pm$1.36     & 87.72$\pm$0.22    & 88.85$\pm$0.22     & 88.25$\pm$0.29     & 87.95$\pm$0.00              \\
                      &                         & $\mathcal{D}_{f}$   & 90.54$\pm$0.00   & 0.00$\pm$0.00       & 0.00$\pm$0.00       & 51.13$\pm$17.37   & 13.95$\pm$6.25     & 0.00$\pm$0.00       & 0.00$\pm$0.00                  \\
                      &                         & MIA    & 80.40$\pm$0.00    & 8.43$\pm$0.02      & 21.97$\pm$0.07     & 0.00$\pm$0.00      & 9.10$\pm$0.05       & 0.77$\pm$0.00      & 3.20$\pm$0.00                \\
                      &                         & $t$    & 549.32$\pm$81.84 & 2176.08$\pm$92.76  & 1524.99$\pm$147.26 & 661.84$\pm$117.05 & 986.10$\pm$149.59   & 1176.24$\pm$136.66 & 767.35$\pm$177.77 \\ \hline
\end{tabular}
\caption{Cifar100 class unlearning. Parameters: ResNet18 $\alpha = 10, \lambda = 1$, Vision Transformer $\alpha = 5, \lambda = 1$; *changing $\alpha$ from 10 to 20 results in [$\mathcal{D}_{r}$ = 72.98, $\mathcal{D}_f$ = 0, MIA = 0.20]. MR: mushrooms, Rkt: rocket. $\mathcal{D}_{r}$ and $\mathcal{D}_f$ report the accuracy on the respective dataset in percent. MIA in percent and time $t$ in seconds.}
\label{tab:appendix}
\end{table*}

\begin{table*}[t!]
\fontsize{9pt}{12pt}\selectfont
\setlength{\tabcolsep}{1.3pt}
\centering
\begin{tabular}{ll|l|c|c|c|c|c|c|c}
\hline
                      & Class                     & metric & baseline           & retrain             & finetune            & teacher          & UNSIR              & amnesiac            & SSD                \\ \hline
\multirow{20}{*}{RN}  & \multirow{4}{*}{ED}       & $\mathcal{D}_{r}$  & 82.56$\pm$0.00    & 82.13$\pm$0.23     & 73.19$\pm$1.08     & 82.04$\pm$0.29    & 80.95$\pm$0.24    & 81.34$\pm$0.18     & 83.15$\pm$0.00    \\
                      &                           & $\mathcal{D}_{f}$   & 82.26$\pm$0.00    & 0.00$\pm$0.00       & 0.00$\pm$0.00       & 10.88$\pm$1.3     & 50.48$\pm$1.87    & 0.00$\pm$0.00       & 1.76$\pm$0.00     \\
                      &                           & MIA    & 89.56$\pm$0.00    & 8.91$\pm$0.01      & 30.37$\pm$0.05     & 0.00$\pm$0.00      & 30.73$\pm$0.04    & 7.02$\pm$0.01      & 4.16$\pm$0.00     \\
                      &                           & $t$    & 109.68$\pm$5.07   & 440.59$\pm$7.71    & 155.44$\pm$5.33    & 117.94$\pm$5.4    & 119.66$\pm$5.6    & 146.56$\pm$6.78    & 120.40$\pm$5.33    \\ \cline{2-10} 
                      & \multirow{4}{*}{NS}       & $\mathcal{D}_{r}$  & 82.10$\pm$0.00     & 81.33$\pm$0.22     & 72.03$\pm$1.63     & 81.36$\pm$0.27    & 80.20$\pm$0.25     & 80.70$\pm$0.4       & 82.33$\pm$0.00    \\
                      &                           & $\mathcal{D}_{f}$   & 91.08$\pm$0.00    & 0.00$\pm$0.00       & 0.00$\pm$0.00       & 10.91$\pm$2.16    & 57.33$\pm$6.0     & 0.00$\pm$0.00       & 0.00$\pm$0.00      \\
                      &                           & MIA    & 88.68$\pm$0.00    & 3.77$\pm$0.01      & 17.60$\pm$0.03      & 0.00$\pm$0.00      & 21.15$\pm$0.07    & 3.71$\pm$0.01      & 3.28$\pm$0.00     \\
                      &                           & $t$    & 113.66$\pm$5.64   & 436.92$\pm$4.59    & 155.93$\pm$5.78    & 119.95$\pm$6.59   & 123.41$\pm$8.03   & 145.75$\pm$7.4     & 123.31$\pm$6.19   \\ \cline{2-10} 
                      & \multirow{4}{*}{people}   & $\mathcal{D}_{r}$  & 82.11$\pm$0.00    & 81.20$\pm$0.19      & 72.46$\pm$1.01     & 81.31$\pm$0.16    & 80.29$\pm$0.23    & 80.64$\pm$0.34     & 82.31$\pm$0.00    \\
                      &                           & $\mathcal{D}_{f}$   & 90.70$\pm$0.00     & 0.00$\pm$0.00       & 0.00$\pm$0.00       & 1.07$\pm$0.48     & 71.36$\pm$3.77    & 0.00$\pm$0.00       & 0.00$\pm$0.00      \\
                      &                           & MIA    & 91.72$\pm$0.00    & 1.36$\pm$0.00      & 16.20$\pm$0.04      & 0.00$\pm$0.00      & 45.20$\pm$0.12     & 6.28$\pm$0.01      & 1.12$\pm$0.00     \\
                      &                           & $t$    & 109.37$\pm$4.12   & 437.96$\pm$5.94    & 154.46$\pm$5.96    & 120.02$\pm$6.22   & 122.50$\pm$7.17    & 147.66$\pm$7.27    & 118.23$\pm$4.78   \\ \cline{2-10} 
                      & \multirow{4}{*}{veg}      & $\mathcal{D}_{r}$  & 82.31$\pm$0.00    & 81.39$\pm$0.21     & 71.42$\pm$1.32     & 81.46$\pm$0.3     & 80.29$\pm$0.26    & 81.01$\pm$0.33     & 82.38$\pm$0.00    \\
                      &                           & $\mathcal{D}_{f}$   & 86.90$\pm$0.00     & 0.00$\pm$0.00       & 0.00$\pm$0.00       & 2.67$\pm$1.35     & 64.45$\pm$1.77    & 0.00$\pm$0.00       & 0.00$\pm$0.00      \\
                      &                           & MIA    & 89.52$\pm$0.00    & 9.74$\pm$0.01      & 29.39$\pm$0.08     & 0.00$\pm$0.00      & 40.66$\pm$0.06    & 5.00$\pm$0.01       & 16.96$\pm$0.00    \\
                      &                           & $t$    & 113.03$\pm$4.85   & 439.21$\pm$7.14    & 154.48$\pm$6.9     & 121.64$\pm$7.17   & 120.38$\pm$5.21   & 147.26$\pm$8.83    & 120.01$\pm$5.11   \\ \cline{2-10} 
                      & \multirow{4}{*}{Veh2} & $\mathcal{D}_{r}$  & 82.69$\pm$0.00    & 82.11$\pm$0.19     & 73.50$\pm$0.86      & 81.96$\pm$0.21    & 80.81$\pm$0.46    & 81.13$\pm$0.3      & 82.97$\pm$0.00    \\
                      &                           & $\mathcal{D}_{f}$   & 80.41$\pm$0.00    & 0.00$\pm$0.00       & 0.00$\pm$0.00       & 3.62$\pm$1.07     & 46.92$\pm$2.27    & 0.00$\pm$0.00       & 0.00$\pm$0.00      \\
                      &                           & MIA    & 82.56$\pm$0.00    & 13.54$\pm$0.01     & 30.63$\pm$0.04     & 0.00$\pm$0.00      & 35.16$\pm$0.03    & 7.54$\pm$0.01      & 6.68$\pm$0.00     \\
                      &                           & $t$    & 113.55$\pm$4.16   & 441.71$\pm$9.48    & 154.60$\pm$3.59     & 120.76$\pm$5.39   & 121.28$\pm$7.24   & 144.03$\pm$5.85    & 121.57$\pm$4.67   \\ \hline \hline
\multirow{20}{*}{ViT} & \multirow{4}{*}{ED}       & $\mathcal{D}_{r}$  & 95.73$\pm$0.00    & 94.71$\pm$0.14     & 87.35$\pm$1.44     & 93.42$\pm$0.62    & 93.25$\pm$0.57    & 93.45$\pm$0.44     & 95.82$\pm$0.00              \\
                      &                           & $\mathcal{D}_{f}$   & 95.03$\pm$0.00    & 0.00$\pm$0.00       & 0.28$\pm$0.22      & 4.14$\pm$3.75     & 73.67$\pm$3.41    & 0.03$\pm$0.08      & 53.53$\pm$0.00              \\
                      &                           & MIA    & 91.60$\pm$0.00     & 9.82$\pm$0.01      & 23.60$\pm$0.04      & 0.02$\pm$0.00     & 38.18$\pm$0.04    & 1.70$\pm$0.00       & 1.32$\pm$0.00               \\
                      &                           & $t$    & 613.47$\pm$76.84  & 2092.64$\pm$110.28 & 1358.86$\pm$66.66  & 712.91$\pm$120.8  & 670.57$\pm$120.82 & 1177.95$\pm$149.07 & 699.88$\pm$80.8   \\ \cline{2-10} 
                      & \multirow{4}{*}{NS}       & $\mathcal{D}_{r}$  & 95.71$\pm$0.00    & 94.79$\pm$0.11     & 87.45$\pm$1.2      & 93.50$\pm$0.4      & 93.09$\pm$0.44    & 93.68$\pm$0.56     & 93.63$\pm$0.00              \\
                      &                           & $\mathcal{D}_{f}$   & 95.37$\pm$0.00    & 0.00$\pm$0.00       & 0.05$\pm$0.15      & 2.63$\pm$2.08     & 50.65$\pm$6.78    & 0.00$\pm$0.00       & 0.00$\pm$0.00                  \\
                      &                           & MIA    & 85.04$\pm$0.00    & 4.70$\pm$0.01       & 16.97$\pm$0.04     & 0.06$\pm$0.00     & 8.58$\pm$0.02     & 1.04$\pm$0.00      & 1.88$\pm$0.00               \\
                      &                           & $t$    & 558.52$\pm$34.14  & 2108.89$\pm$120.12 & 1352.24$\pm$68.52  & 636.55$\pm$76.65  & 596.23$\pm$72.14  & 1117.64$\pm$79.21  & 691.88$\pm$101.13 \\ \cline{2-10} 
                      & \multirow{4}{*}{people}   & $\mathcal{D}_{r}$  & 95.54$\pm$0.00    & 94.54$\pm$0.14     & 80.18$\pm$19.88    & 93.19$\pm$0.54    & 92.74$\pm$0.95    & 93.28$\pm$0.41     & 95.33$\pm$0.00              \\
                      &                           & $\mathcal{D}_{f}$   & 98.54$\pm$0.00    & 0.00$\pm$0.00       & 0.09$\pm$0.14      & 2.91$\pm$3.36     & 93.09$\pm$2.04    & 0.00$\pm$0.00       & 0.00$\pm$0.00                  \\
                      &                           & MIA    & 89.48$\pm$0.00    & 1.56$\pm$0.00      & 8.08$\pm$0.03      & 0.01$\pm$0.00     & 63.42$\pm$0.08    & 0.60$\pm$0.00       & 1.20$\pm$0.00                \\
                      &                           & $t$    & 602.46$\pm$103.93 & 2139.94$\pm$160.1  & 1343.29$\pm$72.31  & 683.55$\pm$106.74 & 658.92$\pm$125.44 & 1147.97$\pm$155.95 & 658.06$\pm$74.03  \\ \cline{2-10} 
                      & \multirow{4}{*}{veg}      & $\mathcal{D}_{r}$  & 95.59$\pm$0.00    & 94.54$\pm$0.21     & 87.09$\pm$1.24     & 92.92$\pm$0.51    & 93.25$\pm$0.35    & 93.29$\pm$0.41     & 95.71$\pm$0.00              \\
                      &                           & $\mathcal{D}_{f}$   & 97.57$\pm$0.00    & 0.00$\pm$0.00       & 0.30$\pm$0.29       & 8.28$\pm$6.79     & 89.02$\pm$2.41    & 0.02$\pm$0.07      & 0.00$\pm$0.00                  \\
                      &                           & MIA    & 91.32$\pm$0.00    & 4.41$\pm$0.01      & 14.72$\pm$0.05     & 0.02$\pm$0.00     & 58.67$\pm$0.04    & 1.02$\pm$0.00      & 1.88$\pm$0.00               \\
                      &                           & $t$    & 610.25$\pm$94.83  & 2108.18$\pm$116.8  & 1443.35$\pm$132.24 & 713.12$\pm$116.66 & 700.74$\pm$99.17  & 1182.43$\pm$134.65 & 755.52$\pm$113.61 \\ \cline{2-10} 
                      & \multirow{4}{*}{Veh2} & $\mathcal{D}_{r}$  & 95.73$\pm$0.00    & 94.85$\pm$0.13     & 87.75$\pm$1.64     & 93.59$\pm$0.3     & 93.56$\pm$0.32    & 93.88$\pm$0.15     & 93.12$\pm$0.00              \\
                      &                           & $\mathcal{D}_{f}$   & 95.22$\pm$0.00    & 0.00$\pm$0.00       & 0.04$\pm$0.12      & 4.88$\pm$4.12     & 70.31$\pm$5.03    & 0.00$\pm$0.00       & 0.00$\pm$0.00                  \\
                      &                           & MIA    & 84.04$\pm$0.00    & 22.96$\pm$0.03     & 38.15$\pm$0.08     & 0.02$\pm$0.00     & 48.98$\pm$0.07    & 1.20$\pm$0.00       & 7.04$\pm$0.00               \\
                      &                           & $t$    & 621.56$\pm$100.35 & 2194.61$\pm$146.18 & 1467.60$\pm$122.95  & 645.57$\pm$78.86  & 718.06$\pm$104.18 & 1216.12$\pm$114.52 & 786.10$\pm$103.29  \\ \hline
\end{tabular}
\caption{Cifar20 class unlearning. Parameters: ResNet18 $\alpha = 10, \lambda = 1$, Vision Transformer $\alpha = 5, \lambda = 1$. ED: electrical devices, NS: natural scenes., Veh2: vehicle2, veg: vegetables; $\mathcal{D}_{r}$ and $\mathcal{D}_f$ report the accuracy on the respective dataset in percent. MIA in percent and time $t$ in seconds.}
\label{tab:appendix}
\end{table*}

\end{document}